\documentclass[conference,a4paper]{IEEEtran}
\usepackage[utf8]{inputenc} 
\usepackage[T1]{fontenc}
\usepackage{url}              
\usepackage{algorithm}
\usepackage{hyperref}
\usepackage{multirow}
\usepackage{bbding}
\usepackage{color}
\usepackage{amsmath,amssymb,amsfonts}
\usepackage{amsthm}
\usepackage{graphicx}
\usepackage{setspace}
\usepackage{multicol}
\usepackage{multirow}
\usepackage{cite}
\usepackage{float}
\usepackage{color}
\usepackage{alltt, listings}
\usepackage{subfigure}
\usepackage{algpseudocode}
\usepackage{wrapfig}
\usepackage{url}
\usepackage{bm}
\usepackage{authblk}

\newtheorem{assumption}{Assumption}
\newtheorem{theorem}{Theorem}
\newtheorem{lemma}{Lemma}

\theoremstyle{definition}

\begin{document}

\title{Truncated Non-Uniform Quantization for Distributed SGD} 

\author[a]{Guangfeng Yan}
\author[b]{Tan Li}
\author[c]{Yuanzhang Xiao}
\author[d]{Congduan Li}
\author[a]{Linqi Song}
\affil[a]{Department of Computer Science, City University of Hong Kong, Hong Kong SAR}
\affil[b]{Department of Computer Science,
The Hang Seng University of Hong Kong, Hong Kong SAR}
\affil[c]{Department of Electrical and Computer Engineering, University of Hawaii at Manoa, United States}
\affil[d]{School of Electronics and Communication Engineering, Sun Yat-sen University, Guangzhou, China}
\renewcommand\Authands{ and }

\maketitle

\begin{abstract}
To address the communication bottleneck challenge in distributed learning, our work introduces a novel two-stage quantization strategy designed to enhance the communication efficiency of distributed Stochastic Gradient Descent (SGD). The proposed method initially employs truncation to mitigate the impact of long-tail noise, followed by a non-uniform quantization of the post-truncation gradients based on their statistical characteristics. We provide a comprehensive convergence analysis of the quantized distributed SGD, establishing theoretical guarantees for its performance. Furthermore, by minimizing the convergence error, we derive optimal closed-form solutions for the truncation threshold and non-uniform quantization levels under given communication constraints. Both theoretical insights and extensive experimental evaluations demonstrate that our proposed algorithm outperforms existing quantization schemes, striking a superior balance between communication efficiency and convergence performance. 
\end{abstract}


\section{Introduction}
\label{introduction}
Distributed Stochastic Gradient Descent (DSGD)~\cite{{dean2012large},{bekkerman2011scaling}} is a popular algorithm that utilizes local client data to build distributed learning models. In DSGD, local gradients are computed at each client and transferred to a parameter server, which aggregates them to update the global model. This process is repeated until all nodes reach a global consensus on the learning model. To alleviate the communication bottleneck in distributed learning, various model compression techniques have been applied to local gradients to enhance communication efficiency. Sparsification and quantization are among the most widely adopted strategies. 

Among the prevalent quantization methods, uniform quantizers~\cite{alistarh2017qsgd} are commonly utilized due to their simplicity and ease of implementation. However, this approach does not adequately represent the typical bell-shaped with a long-tailed distribution of weights and activations in neural networks~\cite{wen2017terngrad}. A natural approach to managing gradients with a long-tail distribution is to implement gradient truncation~\cite{banner2019post,chen2023quantizing}, which involves establishing a threshold that serves to mitigate the impact of extreme gradient values on quantization. Another method is to design a non-uniform quantizer~\cite{panter1951quantization, algazi1966useful}. Most of the work in these two areas largely stems from empirical engineering practices, such as manually setting clipping thresholds or assigning more quantization points to areas of high data density—the peaks of the distribution—and fewer points within the less dense tails.

Only a handful of studies attempt to theoretically guide the designing of quantizers, achieving limited success. For example,~\cite{banner2019post} assumes a Laplace distribution for the gradients to find the optimal truncation threshold, but then applies a simple uniform quantizer for the compression; while another study~\cite{chen2023nqfl}, introduces a non-uniform quantizer based on the Lloyd-Max algorithm~\cite{lloyd1982least} in the federated learning setting. However, this method is computationally intensive and does not always yield an optimal solution. A framework to jointly optimize the truncation threshold and non-uniform quantizer parameters is lacking. 

Our work aims to fill this gap. We notice that truncation and non-uniform quantization have been extensively used in signal processing for wireless communications. Yet, the goal is mainly to minimize the signal quantization distortion, e.g., using mean square errors. When being used in distributed learning to train a deep neural network model jointly, the goal is to reduce the speed degradation of the learning convergence. We innovatively combine gradient truncation with non-uniform quantization and extend this hybrid approach to a distributed learning framework with the aim of minimizing the convergence speed. Starting from this, we design our truncation and non-uniform quantization schemes. We address the joint optimization challenge and provide a solid theoretic analysis for communication and convergence trade-offs, in Laplace distribution assumptions, which was shown to be a close approximation to the gradient distribution in deep neural networks~\cite{banner2019post}. The main contribution of this work is summarized as follows. 

$\bullet$ We design a novel truncated non-uniform quantizer and integrate it into a distributed SGD framework under communication constraints.

$\bullet$ We provide a theoretical framework for analyzing the impact of the designed quantizer on convergence error.

$\bullet$ We derive optimal closed-form solutions for the with the assumption of Laplace gradient distribution.

$\bullet$ Both theoretical and numerical evaluations show that our proposed method outperforms the benchmarks and is even competitive with the non-compressed models.

\section{Problem Formulation}

We consider a distributed learning problem with $N$ clients and a central server. These $N$ clients collaboratively train a global model through communication with the central server via a master-worker manner. Each client $i$ has some local data, denoted by $\mathcal{D}^{(i)}$. 
The objective is to minimize the empirical risk over all the local data across clients, i.e., solve the optimization problem 
\begin{equation}
\begin{array}{llll}	\min_{\bm{\theta} \in \mathbb{R}^d}F(\bm{\theta}) = \sum_{i=1}^N  w_i \mathbb{E}_{\xi^{(i)}\sim \mathcal{D}^{(i)}}\left[\ell(\bm{\theta};\xi^{(i)})\right],
	\end{array}\label{optim_problem}
\end{equation}
where $w_i = \frac{|\mathcal{D}^{(i)}|}{\sum_{i=1}^N|\mathcal{D}^{(i)}|}$ is the weight of client $i$, $\xi^{(i)}$ is a random sample from $\mathcal{D}^{(i)}$ and $\ell(\bm{\theta};\xi^{(i)})$ is the local loss function of the model $\bm{\theta}$ at one data sample $\xi^{(i)}$. 

Distributed SGD is a widely used approach to solve this problem, especially for deep neural networks~\cite{{yan2022ac},{bekkerman2011scaling}}. In this setting, each client $i$ first downloads the global model $\bm{\theta}_t$ from the server at iteration $t$, then randomly selects a batch of samples $B^{(i)}_t \subseteq D^{(i)}$ of size $B$ to compute its local stochastic gradient with respect to $\bm{\theta}_t$: $\bm{g}^{(i)}_t = \frac{1}{B} \sum_{\xi^{(i)} \in B^{(i)}_t} \nabla \ell(\bm{\theta}_t;\xi^{(i)})$. Then the server aggregates these gradients and updates the model: $\bm{\theta}_{t+1} = \bm{\theta}_t - \eta\sum_{i=1}^N w_i \bm{g}^{(i)}_t$, where $\eta$ is the server learning rate. 
 To reduce the communication cost, we compress the local stochastic gradients before sending them to the server: $\bm{\theta}_{t+1} = \bm{\theta}_t - \eta\sum_{i=1}^N w_i \mathcal{C}_{b}[\bm{g}^{(i)}_t]$, 
where $\mathcal{C}_{b}[\cdot]$ is the compressor operator to compress each element of $\bm{g}^{(i)}_t$ into $b$ bits. 

In this paper, we aim to design a general non-uniform compressor using a two-stage quantizer to solve the distributed SGD problem given communication constraints.

\section{Truncated Non-Uniform Quantizer for Distributed SGD}
In this section, we introduce a two-stage quantizer that combines truncation with non-uniform quantization. Following this, we incorporate the quantizer into a distributed SGD algorithm and present an analysis of its convergence error.

\textbf{Gradient Truncation} The truncation operation cuts off the gradient so that the value is within a range. For an element $g$ of gradient $\bm{g}$, the $\alpha$-truncated operator $\mathcal{T}_{\alpha}[g]$ is defined as 
\begin{equation}
     \mathcal{T}_{\alpha}[g] = \begin{cases} g,& \text{for $|g|\le \alpha$,}\\ 
    \text{sgn}(g)\cdot \alpha,& \text{for $|g|> \alpha$} \end{cases}
 \label{eq:truncated_operation}
\end{equation} 
where $\alpha$ is a truncation threshold that determines the range of gradients, and $\text{sgn}(g)\in\{+1,-1\}$ is the sign of $g$. A common intuition is that the thicker the tail of the gradient distribution, the larger the value of $\alpha$ should be set to ensure that the discarded gradient information is upper bounded.

\textbf{Nonuniform Gradient Quantization} For the truncated gradient, we propose a novel element-wise non-uniform quantization scheme. Specifically, consider a truncated gradient element $g$ that falls within the interval $[a_1,a_2]$. To satisfy communication constraints, we aim to encode it using $b$ bits. This encoding process results in $2^b$ discrete quantization points, which effectively divide the interval $[a_1,a_2]$ into $s=2^b-1$ disjoint intervals. The boundaries of these intervals are defined by the points $a_1=l_0 < l_1 \ldots < l_s=a_2$. Each $k$-th interval is denoted by $\Delta_k \triangleq [l_{k-1}, l_k]$, and has a length (or a quantization step size) of $|\Delta_k| = l_k - l_{k-1}$. If $g \in \Delta_k$, we have
\begin{equation}\
	\mathcal{Q}[g] = \begin{cases} l_{k-1},& \text{with probability $1-p_r$,}\\ 
	l_k,& \text{with probability $p_r = \cfrac{g-l_{k-1}}{|\Delta_k|}$.} \end{cases}
\label{eq:quantized_operation}
\end{equation} 
It is evident that the specific operation of the quantizer depends on the quantization step size $\Delta_k$, which is essentially the quantization points $\mathcal{L}\triangleq\{l_0,l_1,...,l_s\}$. 

We make the following two assumptions on the raw gradient $\nabla l(\bm{\theta}_t;\xi^{(i)})$ and the objective function $F(\bm{\theta})$~\cite{{bottou2018optimization},{data2023byzantine}}:
\begin{assumption}[Bounded Variance]
	For parameter $\bm{\theta}_t$, the stochastic gradient $\nabla \ell(\bm{\theta}_t;\xi^{(i)})$ sampled from any local dataset have uniformly bounded variance for all clients:
	\begin{align}
	\mathbb{E}_{\xi^{(i)}\sim \mathcal{D}^{(i)}}\left[\|\nabla \ell(\bm{\theta}_t;\xi^{(i)})-\nabla F(\bm{\theta}_t)\|^2\right] \le \sigma^2.
	\end{align}
	\label{ass:stochastic_gradient} 
\end{assumption}
\vspace{-5mm}
\begin{assumption}[Smoothness]
	The objective function $F(\bm{\theta})$ is $\nu$-smooth: $\forall \bm{\theta},\bm{\theta}' \in \mathbb{R}^d$, $\|\nabla F(\bm{\theta})-\nabla F(\bm{\theta}')\| \leq \nu\|\bm{\theta}-\bm{\theta}'\|$.
	\label{ass:smoothnesee} 
\end{assumption}

This also determines the statistical characteristics of the quantizer, as demonstrated by the following lemma.

\begin{lemma}[Unbiasness and Bounded Variance]
    For a truncated gradient element $g\in [a_1,a_2]$ with probability density function $p_g(\cdot)$, given the quantization points $\mathcal{L}=\{l_0, l_1,...,l_s\}$, the nonuniform stochastic quantization satisfies:
    \begin{equation}
		\mathbb{E}[\mathcal{Q}[g]] = g
		\label{eq:unbiassness}
	\end{equation}
	and
	\begin{equation}
	  \mathbb{E} \|\mathcal{Q}[g] - g\|^2 \le \sum_{k=1}^{s} \frac{P_k|\Delta_k|^2}{4}
	\label{eq:qsg}
	\end{equation}
	where $P_k = \int_{l_{k-1}}^{l_k}p_g(x)\mathrm{d}x$ and $|\Delta_k|  = l_k - l_{k-1}$.
	\label{lem:qsg}    
\end{lemma}
The complete proof can be found in Appendix~\ref{proof_lemma1}. We further introduce the concept of the ``density'' of quantization points, defined as $\lambda_s(g) \triangleq \frac{1}{|\Delta(g)|}$. This definition ensures that $\int_{a_1}^{a_2} \lambda_s(g) dg = s$. In the remainder of the paper, we denote a non-uniform quantizer with quantization destiny function$\lambda_s(\cdot)$ by $Q_{\lambda_s}[\cdot]$. By doing this, Lemma~\ref{lem:qsg} can be rewritten as $\mathbb{E}[\mathcal{Q}_{\lambda_s}[g]] = g$ and $\mathbb{E} \|\mathcal{Q}_{\lambda_s}[g] - g\|^2 \le \int_{a_1}^{a_2}\frac{p(g)}{4\lambda_s(g)^2}\mathrm{d}g$. In a specific instance, if we take $\lambda_s(g) = \frac{s}{a_2-a_1}$, then the nonuniform quantization simplifies to uniform quantization~\cite{alistarh2017qsgd}, i.e., $\mathcal{L}=\{a_1 + k \frac{a_2-a_1}{s}, k=0,1,...,s\}$.

\begin{figure}[ht!]
\centerline{\includegraphics[width=0.85\linewidth]{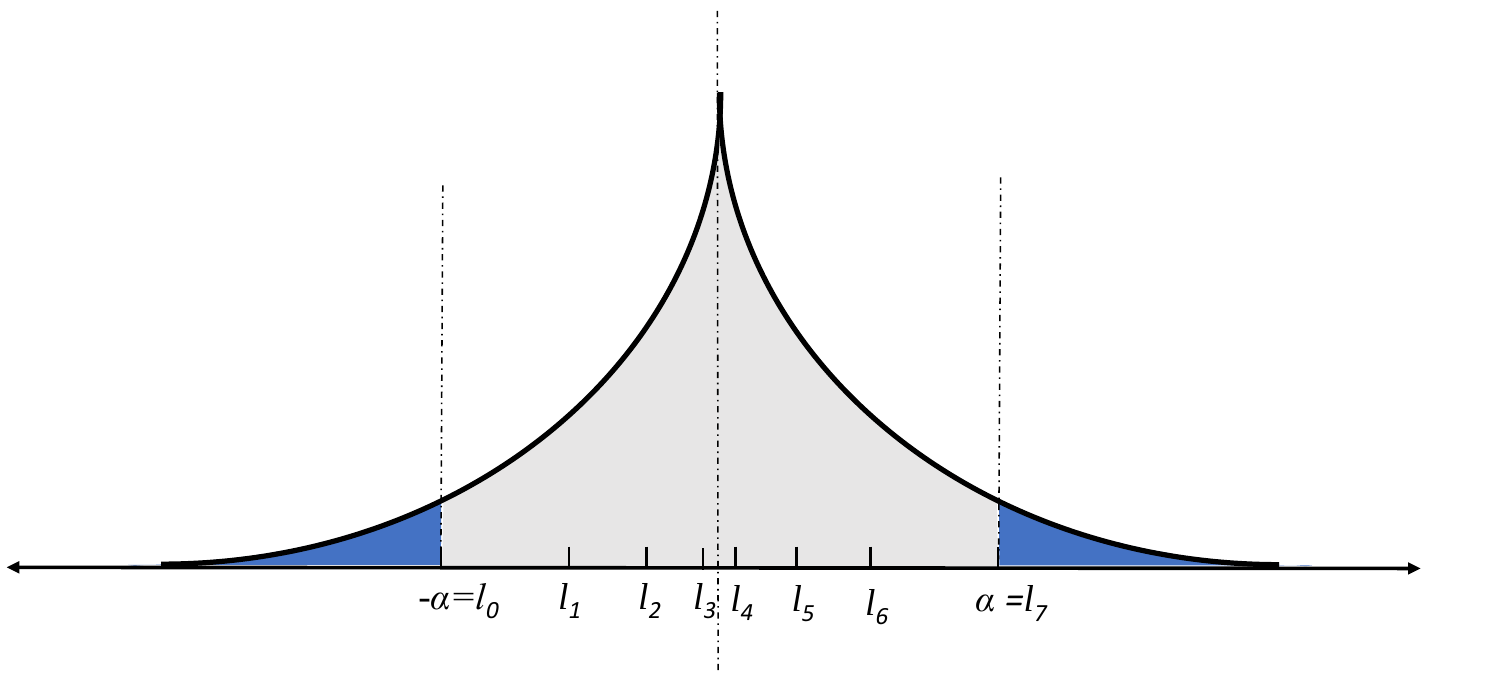}}
	\caption{Truncated Non-Uniform Quantizer (With truncation threshold $[-\alpha, \alpha]$ and quantization bit $b=3$ and quantization level $s=7$.)}
	\label{fig:Truncated_Quantization}
\end{figure}

To summarize, our proposed truncated non-uniform quantizer, denoted as $\mathcal{Q}_{\lambda_s}[\mathcal{T}_{\alpha}(\bm{g})]$, begins with the truncation of gradients $\bm{g}$ using $\mathcal{T}_{\alpha}(\bm{g})$ to curtail values outside the $[-\alpha, \alpha]$ range, thereby reducing noise. These truncated gradients are then quantized through $\mathcal{Q}_{\lambda_s}[\cdot]$ into $b$-bit representations $(b = log_2 (s+1))$, as depicted in Fig.~\ref{fig:Truncated_Quantization}. The entire process is encapsulated in the Truncated Non-uniform Quantization for Distributed SGD (TNQSGD) algorithm, detailed in Alg.~\ref{alg:TQSGD}, which integrates our quantization method into the distributed SGD framework to enhance communication efficiency without compromising convergence performance.

\begin{algorithm}[ht] 
	\caption{Truncated Non-uniform Quantization for Distributed SGD (TNQSGD)}
	\begin{algorithmic}[1]
	\State \textbf{Input:} Learning rate $\eta$, initial point $\bm{\theta}_0 \in \mathbb{R}^d$, communication round $T$, parameters of $\mathcal{Q}_{\lambda_s}[\mathcal{T}_{\alpha}(\cdot)]$ (truncated threshold $\alpha$, quantization density function $\lambda_s$);
		\For {each communication rounds $t = 0, 1, ..., T-1$:}
		\State \textbf{On each client {$i=1, ..., N$}:}
		\State Download $\bm{\theta}_t $ from server;
		\State Compute the local gradient $\bm{g}^{(i)}_t$ using SGD;
        \State Quantize $\bm{g}^{(i)}_t$ to $\bm{\hat{g}}^{(i)}_t = \mathcal{Q}_{\lambda_s}[\mathcal{T}_{\alpha}(\bm{g}^{(i)}_t)]$ using Eq.~\eqref{eq:truncated_operation} and~\eqref{eq:quantized_operation};
		\State Send $\bm{\hat{g}}^{(i)}_t$ to the server;
 	\State \textbf{On the server:}
        \State Aggregate all quantized gradients  $\bm{\bar g}_t = \sum_{i=1}^N w_i \bm{\hat{g}}^{(i)}_t$;
		\State Update global model parameter: $\bm{\theta}_{t+1} = \bm{\theta}_t - \eta \bm{\bar g}_t$;
	\EndFor
\end{algorithmic} 
	\label{alg:TQSGD}
\end{algorithm}
 
Assuming that each element follows a symmetrical probability density around zero $p(g)$ and is independently and identically distributed, we have the following Lemma to characterize the convergence performance of TNQSGD.

\begin{lemma}
    For a $N$-client distributed learning problem, by applying the quantizer $\mathcal{Q}_{\lambda_s}[\mathcal{T}_{\alpha}(\cdot)]$ and $w_i = \frac{1}{N}$, the convergence error of Alg.~\ref{alg:TQSGD} for the smooth objective is upper bounded by
	\begin{align} \label{eq:cov_of_TQSGD}
		&\frac{1}{T}\sum_{t=0}^{T-1} \|\nabla F(\bm{\theta}_t)\|^2 \le \underbrace{\frac{2[F(\bm{\theta}_0)-F(\bm{\theta}^*)]}{T\eta}+ \frac{\sigma^2}{NB}}_{\triangleq\mathcal{E}_{DSGD}} \nonumber\\
        &~~+\underbrace{\cfrac{d}{4N}\int_{-\alpha}^{\alpha}\frac{p(g)}{\lambda_s(g)^2}\mathrm{d}g + \cfrac{2d}{N}\int_{\alpha}^{+\infty}(g-\alpha)^2p(g)\mathrm{d}g}_{\triangleq\mathcal{E}_{TQ}} 
	\end{align}
\end{lemma}
Several insights can be derived from this Lemma. The term labeled $\mathcal{E}_{DSGD}$ in Eq.~\ref{eq:cov_of_TQSGD} represents the bound on the convergence error for the standard distributed SGD when applied with \textit{non-compressed} model updates. The subsequent term, $\mathcal{E}_{TQ}$, encapsulates the error introduced by our proposed two-stage quantizer, which highlights the balance between the level of compression and the precision of our proposed algorithm. The term $\mathcal{E}_{TQ}$ can be further broken down into two components: a variance component due to quantization  (the first term) and a bias component due to the truncation (the second term). It is important to note that with a sufficiently small truncation threshold $\alpha$, the density of quantization points described by $\lambda_s(x)$ will be notably high. This concentration results in a reduction of the quantization variance towards zero, whilst simultaneously increasing the truncation bias. Conversely, as the threshold $\alpha$ increases, the truncation bias diminishes towards zero, but this leads to a rise in quantization variance. Additionally, the distribution of quantization points $\lambda_s(g)$ has a direct impact on the level of quantization variance, and thus influences the overall value of $\mathcal{E}_{TQ}$. For a detailed proof, refer to the Appendix~\ref{proof_lemma2}.

\section{Optimal Parameters Design For Truncated Non-Uniform Quantizer}
In this section, we aim to provide theoretical guidance for optimizing the parameters of the proposed quantizer. This can be addressed by solving a joint optimization problem involving two parameters, the truncation threshold and the quantization parameter.
\subsection{Optimal Parameters for Any Gradient Distribution}
Formally, we formulate the parameter selection problem  as a \textit{convergence error minimization} problem  under the communication constraints: 
\begin{align}
& \min_{\alpha, \lambda_s} ~ \mathcal{E}_{TQ}(\alpha, \lambda_s) \nonumber\\
& s.t. ~~ \int_{-\alpha}^{\alpha} \lambda_s(x) dx = s
\label{eq:DQP}   
\end{align}

From Eq.~\eqref{eq:cov_of_TQSGD}, we find that only the first term of $\mathcal{E}_{TQ}$ contains the quantization density function $\lambda_s(g)$. Hence we investigate solution of $\lambda_s(g)$ by constructing the following Lagrange equation ~\cite{gelfand2000calculus} using the variational principle:
\begin{align}
    I(\lambda_s(g), \nu) = \int_{-\alpha}^{\alpha}\Big[\cfrac{p(g)}{\lambda_s(g)^2} - \mu \lambda_s(g)\Big] \mathrm{d}g
\end{align}
To solve $\lambda_s(g)$ by applying the Euler-Lagrange equation:
\begin{align}
	-\cfrac{2p(g)}{\lambda_s(g)^3}  - \mu = 0
\end{align}

We can obtain $\lambda_s(g) = -(\frac{2p(g)}{\mu})^{\frac{1}{3}}$. 
Further using the communication budget constraints Eq.~\eqref{eq:DQP}:
\begin{align}\label{eq:nonuiform_lambda}
 { \lambda_s(g) = \cfrac{p(g)^{\frac{1}{3}}}{\int_{-\alpha}^{\alpha} p(g)^{\frac{1}{3}} \mathrm{d}g} \cdot s}
\end{align}
From the above expression, we can derive some insights. Given fixed values of $s$ and $\alpha$, a larger $p(g)$ necessitates more quantization bits for effective compression. For a given gradient distribution $p(g)$ and communication constraint $s$, a larger truncation threshold $\alpha$ means retaining larger quantization range. As reflected in Eq.~\eqref{eq:nonuiform_lambda}, this would increase the numerator, thereby decreasing the quantization points density. 

Substitute Eq.~\eqref{eq:nonuiform_lambda} into Eq.~\eqref{eq:cov_of_TQSGD}, $\mathcal{E}_{TQ}$ can be rewrited as:

\begin{align}\label{eq:error_TQ}
    \mathcal{E}_{TQ}(\alpha) = \cfrac{d}{4Ns^2}\Big[\int_{-\alpha}^{\alpha} p(g)^{\frac{1}{3}} \mathrm{d}g\Big]^3 + \cfrac{2d}{N}\int_{\alpha}^{+\infty}(g-\alpha)^2p(g)\mathrm{d}g
\end{align}

The optimum $\alpha$ can be found:
\begin{align}
    \alpha^* = \mathop{\arg\min}_{\alpha} \mathcal{E}_{TQ}(\alpha)
\end{align}
To examine the specific form of $\alpha$, it is necessary to make an assumption about the $p(g)$, i.e., the distribution of the gradients.~\cite{wen2017terngrad} demonstrates that the distribution of coordinates in each local gradient vector typically exhibits a bell-curve shape, akin to that of a Gaussian or Laplace distribution. In our work, we opt for the Laplace distribution which has heavier tails than the Gaussian distribution. This property allows us to capture the potentially larger deviations that are often present in the gradient values.
\subsection{Optimal Parameters for Laplace Gradient Distribution}
The Laplace gradient distribution is defined by:
\begin{align}\label{eq:laplace_gradient}
    p(g|\gamma) = Laplace(g|0, \gamma) = \cfrac{1}{2\gamma} \exp{\{-\frac{|g|}{\gamma}\}}
\end{align}
where $\gamma$ is a scale parameter that indicates the range of variation in gradient values. A larger $\gamma$  suggests a wider variation in gradient values, with heavier tails in the distribution.

Using the Laplace gradient distribution, Eq.~\eqref{eq:error_TQ} can be rewritten as:
\begin{align}\label{eq:error_NQ}
   \mathcal{E}_{TQ}(\alpha) = \cfrac{27d\gamma^2}{Ns^2} \{1-\exp{[\frac{-\alpha}{3\gamma}]}\}^3 + \cfrac{2d\gamma^2}{N}\exp{[-\frac{\alpha}{\gamma}]}
\end{align}

We can get $\alpha$ by minimizing Eq.~\eqref{eq:error_NQ}:
\begin{equation}\label{eq:nonuniform_alpha}
	{\boxed {\alpha = 3\ln{\Big[1+\frac{\sqrt{6}s}{9}\Big]}\gamma}}
\end{equation}
The selection of quantization bit $b= 2, 3, 4$ (i.e., $s=3, 7, 15$) leads to $\alpha = 1.79\gamma, 3.20\gamma, 4.88\gamma$, and $\mathcal{E}_{TQ} = \frac{0.61d\gamma^2}{N}, \frac{0.24d\gamma^2}{N}, \frac{0.077d\gamma^2}{N}$, respectively. Substituting Eq.~\eqref{eq:nonuniform_alpha} into Eq.~\eqref{eq:nonuiform_lambda} and using $p(g) = Laplace(g|0, \gamma)$, we can get 
\begin{align}\label{eq:nonuiform_lambda2}
		{\boxed {\lambda_s(g) = \cfrac{3\sqrt{6}+2s}{8\gamma} \exp{[\frac{-|g|}{3\gamma}]}}}
\end{align}
Eq.~\eqref{eq:nonuniform_alpha} and Eq.~\eqref{eq:nonuiform_lambda2} represent the optimal solutions for the parameters of our designed two-stage quantizer. We next substitute them back into Eq.~\eqref{eq:cov_of_TQSGD} to examine the minimum convergence error that can be guaranteed with this set of parameters. The result is shown in the following Theorem.


\begin{theorem}
    For a $N$-client distributed learning problem with constrained quantization level $s$, using $\alpha$ in Eq.~\eqref{eq:nonuniform_alpha} and $\lambda_s(g)$ in Eq.~\eqref{eq:nonuiform_lambda2}, the convergence error of Alg.~\ref{alg:TQSGD} for the smooth objective is upper bounded by
	\begin{align}\label{eq:convergence_TNQ}
		&\frac{1}{T}\sum_{t=0}^{T-1} \|\nabla F(\bm{\theta}_t)\|^2 \le \mathcal{E}_{DSGD} + \underbrace{\cfrac{27d\gamma^2}{N(s+\frac{3\sqrt{6}}{2})^2}}_{\mathcal{E}_{TQ}} 
	\end{align}
\end{theorem}

We can more intuitively observe how the communication constraint and gradient distribution affect model performance. First, there is a clear trade-off between the available communication resources (number of bits) and model convergence. To elaborate, when the communication constraint is more stringent, the number of bits available for communication is less (less $s$), which leads to a loss of information and poorer performance. Additionally, a larger $\gamma$ suggests a wider variation in gradient values, with heavier tails in the distribution, which also hurts convergence. The complete proof can be found in Appendix~\ref{proof_theorem1}.

\subsection{Relationship With Other Quantization Scheme}

We next examine the effects of truncation and quantization on gradient compression by considering three comparative scenarios based on whether truncation is applied and the type of quantization used—uniform or non-uniform. The scenarios are as follows:

\textbf{Non-uniform Quantization without Truncation}. In this case, we apply NSQ directly to the gradients without any preliminary truncation step, i.e., let $\alpha = \|\bm{g}\|_{\infty}\triangleq \max_j |g_j|$. We use the following lemma to character the upper bound of $\alpha$, i.e., $\|\bm{g}\|_{\infty}$. 

\begin{lemma}\label{lemma:gradients_norm}
    If the gradients follow Laplace distribution $Laplace(g|0, \gamma)$, $\|\bm{g}\|_{\infty}$ satisfied:
    \begin{align}
        \mathbb{E}[\|\bm{g}\|_{\infty}^2] \le 4 \gamma^2 [\ln{2d}]^2
    \end{align}
\end{lemma}
The complete proof can be found in Appendix~\ref{proof_lemma3}.
Using Lemma~\ref{lemma:gradients_norm} and Eq.~\eqref{eq:error_NQ}, the error introduced by non-uniform quantization without truncation can be rewritten as:
\begin{align}\label{eq:convergence_NQ}
   \mathcal{E}_{TQ}^{N} \le \cfrac{27d\gamma^2}{Ns^2}
\end{align}
The selection of quantization bit $b= 2, 3, 4$ leads to $\mathcal{E}_{TQ}^{N} = \frac{3d\gamma^2}{N}, \frac{0.55d\gamma^2}{N}, \frac{0.12d\gamma^2}{N}$, respectively. Comparing equation Eq.~\eqref{eq:convergence_TNQ} and Eq.~\eqref{eq:convergence_NQ}, we find that only using non-uniform quantization without truncation incurs a larger quantization error.

\textbf{Uniform Quantization with Truncation}. Then, we apply truncation on gradients before being quantized uniformly. Truncation can help in reducing the range of gradient values to be quantized, which may lead to a more efficient uniform quantization process. Uniform quantization assigns equal-sized intervals for all values. That is, we set $\lambda_s(g) = \frac{s}{2\alpha}$, which is a typical uniform quantizer. Then $\mathcal{E}_{TQ}$ in Eq.~\eqref{eq:cov_of_TQSGD} can be rewritten as:
\begin{align}\label{eq:error_UQ}
   \mathcal{E}_{TQ}^{UT}(\alpha) = \cfrac{d\alpha^2}{Ns^2} + \cfrac{2d\gamma^2}{N}\exp{[-\frac{\alpha}{\gamma}]}
\end{align}
Hence, we can obtain the optimal solution of $\alpha$ by minimizing Eq.~\eqref{eq:error_UQ}:
\begin{equation}\label{eq:uniform_alpha}
	{\boxed {\alpha = v(s)\gamma}}
\end{equation}
where $v(s)$ satisfied $v\exp{[v]}=s^2$. The numerical result for bits $b= 2, 3, 4$ results with $\alpha = 1.68\gamma, 2.85\gamma, 4.02\gamma$, and $\mathcal{E}_{TQ}^{UT} = \frac{0.69d\gamma^2}{N}, \frac{0.28d\gamma^2}{N}, \frac{0.11d\gamma^2}{N}$, respectively. Substituting $\alpha$ in Eq.~\eqref{eq:uniform_alpha} to Eq.~\eqref{eq:error_UQ}, we can derive the quantization error in this case as: 
\begin{align}
   \mathcal{E}_{TQ}^{UT} &= \cfrac{2d\gamma^2}{3Ns^2} [v(s)^2 + v(s)]
\end{align}

 Compared to the results of truncation with non-uniform quantization (Theorem 1), we can see that we need to set a larger truncation threshold, which will result in a larger quantization error. 

\textbf{Uniform Quantization without Truncation}. In this scenario, we apply uniform quantization to the full range of gradient values without any truncation. Using Lemma~\ref{lemma:gradients_norm} and Eq.~\eqref{eq:error_UQ}, the quantization error can be derived as:
\begin{align}\label{eq:convergence_UQ}
   \mathcal{E}_{TQ}^{U} \le \cfrac{4d\gamma^2}{Ns^2}(\ln{2d})^2
\end{align}

For example, take $d = 5\times10^5$ and $b= 2, 3, 4$, then $\mathcal{E}_{TQ}^{U} = \frac{84.83d\gamma^2}{N}, \frac{15.58d\gamma^2}{N}, \frac{3.39d\gamma^2}{N}$, respectively. We can see that using only uniform quantization without any truncation will result in errors far larger than the other three cases.
\section{Experiments}

In this section, we conduct experiments on MNIST to empirically validate our proposed TNQSGD. 
We compare our methods with three baselines discussed in the previous section: 1) \textbf{QSGD}~\cite{alistarh2017qsgd}: uniform quantization without truncation
operation; 2) \textbf{NQSGD}: non-uniform quantization without truncation
operation;  3) \textbf{TQSGD}: truncation and uniform quantization; and 4) oracle \textbf{DSGD} with clients sending non-compressed gradients to the server.

\textbf{Experimental Setting.}  We conduct experiments for $N = 8$ clients and use AlexNet~\cite{krizhevsky2012imagenet} for all clients. We select the momentum SGD as an optimizer, where the learning rate is set to 0.01, the momentum is set to 0.9, and weight decay is set to 0.0005. Note that gradients from convolutional layers and fully-connected layers have different distributions~\cite{wen2017terngrad}.
We thus quantize convolutional layers and fully-connected layers independently. We estimate $\gamma$ based on maximum likelihood estimation: $\gamma = \frac{\sum_{j=1}^d|g_j|}{d}$.

\begin{figure}
    \centering
    \includegraphics[width=0.75\linewidth]{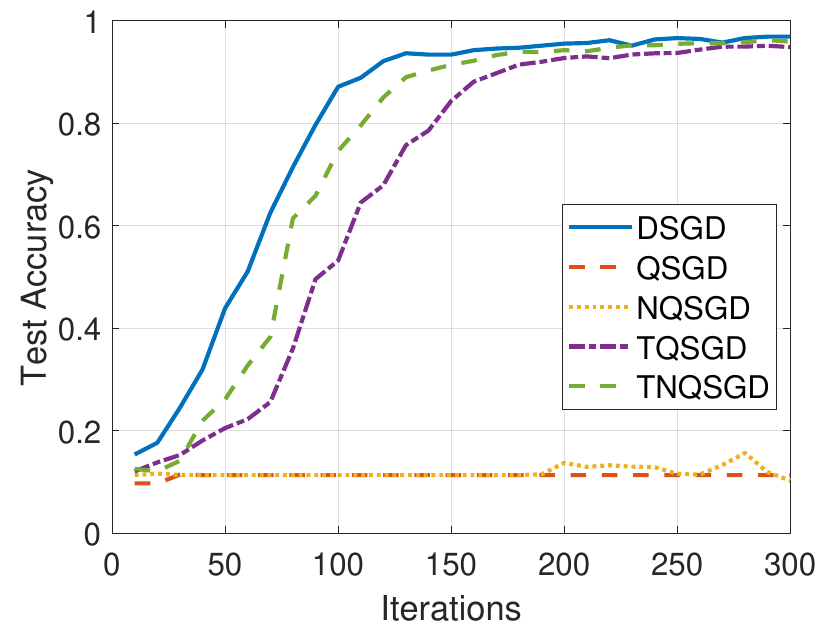}
    \caption{Model performance of different algorithms.}
    \label{fig:testing_performance}
\end{figure}
Fig.~\ref{fig:testing_performance} illustrates the test accuracy of algorithms on MNIST. DSGD achieves a test accuracy of 0.9691 with 32-bit full precision gradients. When $b=3$ bits, TUQSGD and TNQSGD achieve test accuracies of 0.9487 and 0.9595, respectively. In contrast, QSGD and NQSGD are almost unable to converge. The results demonstrate that truncation operation can significantly improve the test accuracy of the model under the same communication constraints. Additionally, non-uniform quantization can further enhance the performance of the algorithm.

\begin{figure}
    \centering
    \includegraphics[width=0.75\linewidth]{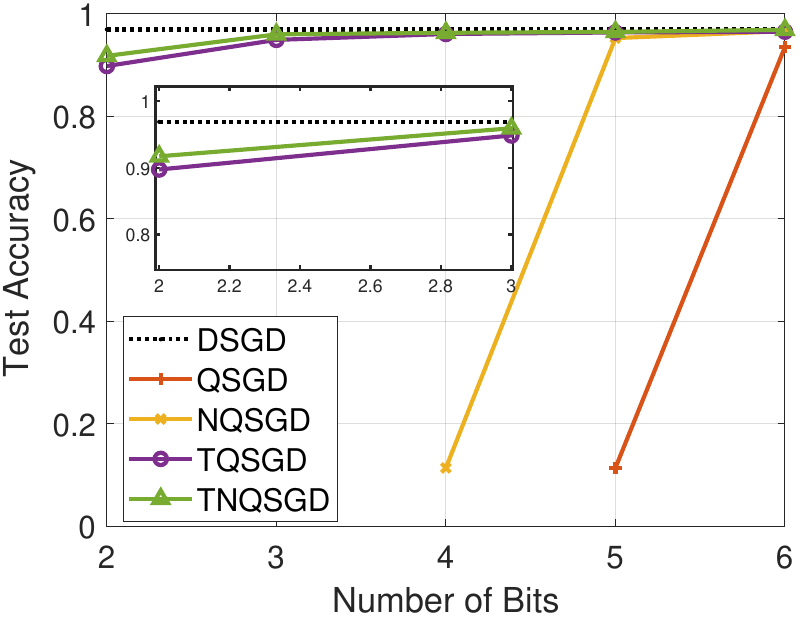}
    \caption{Communication-learning tradeoff of different algorithms.}
    \label{fig:Communication-Learning_tradeoff}
\end{figure}

Fig.~\ref{fig:Communication-Learning_tradeoff} illustrates the tradeoff between communication budget and learning performance in terms of test accuracy of various algorithms. We compare this tradeoff between our proposed algorithms and two other baselines - QSGD and NQSGD. Additionally, we list the accuracy achieved by DSGD without communication budget constraints as a benchmark. All four algorithms exhibit a communication-learning tradeoff; that is, the higher the communication budget, the higher the test accuracy. However, our proposed TUQSGD and TNQSGD achieve higher test accuracies than the other two under the same communication cost.
\section{Conclusion}
In conclusion, in distributed learning, our work presents a two-stage quantization strategy that incorporates an initial truncation step to reduce the influence of long-tail noise, succeeded by a non-uniform quantization tailored to the statistical properties of the truncated gradients. 
By optimizing the convergence error, we have formulated optimal closed-form solutions for setting the truncation threshold and non-uniform quantization parameters within specified communication constraints. Our findings and experimental results confirm that the proposed quantization approach surpasses existing methods, achieving superior convergence performance under equivalent communication constraints.

\newpage
\bibliographystyle{IEEEtran}
\bibliography{mybib}
\clearpage
\section{Appendix}
\subsection{Proof of Lemma 1}
\label{proof_lemma1}
If $x \in \Delta_k$, we have
\begin{align*}\
	\mathcal{Q}[x] = \begin{cases} l_{k-1},& \text{with probability $1-p_r$,}\\ 
	l_k,& \text{with probability $p_r = \cfrac{x-l_{k-1}}{|\Delta_k|}$.} \end{cases}
\end{align*} 

Hence,



\begin{align*}
	 &~~~~~\mathbb{E} \|\mathcal{Q}[x] - x\|^2\\
  &= \sum_{k=1}^{s} \int_{l_{k-1}}^{l_k} \Big[ (l_{k-1}-x)^2\cfrac{l_k-x}{|\Delta_k|} + (l_k-x)^2\cfrac{x-l_{k-1}}{|\Delta_k|}\Big]p(x)\mathrm{d}x\\
  &= \sum_{k=1}^{s}|\Delta_k|^2 \int_{l_{k-1}}^{l_k} \Big[ \cfrac{l_k-x}{|\Delta_k|}\cdot\cfrac{x-l_{k-1}}{|\Delta_k|} \Big]p(x)\mathrm{d}x\\
  &\overset{(b)}{\le} \sum_{k=1}^{s}|\Delta_k|^2 \int_{l_{k-1}}^{l_k} \cfrac{p(x)}{4}\mathrm{d}x\\
  & = \sum_{k=1}^{s} P_k \cfrac{|\Delta_k|^2}{4}
\end{align*}
where $(b)$ uses $y(1-y)\le\cfrac{1}{4}$ for all $y\in[0,1]$, and $P_k = \int_{l_{k-1}}^{l_k}p(x)\mathrm{d}x$.

\subsection{Proof of Lemma 2}
\label{proof_lemma2}

Firstly, we can decompose the mean squared error of the compressed gradient $\mathcal{Q}_{\lambda_s}[ \mathcal{T}_{\alpha}(\bm{g})]$ into a variance term (due to the
nonuniform quantization) and a bias term (due to the truncated operation):

\begin{align}\label{eq:bias_varance}
    \mathbb{E}[\|\mathcal{Q}_{\lambda_s}[ \mathcal{T}_{\alpha}(\bm{g})] - \bm{g}\|^2] &= \underbrace{d\int_{-\alpha}^{\alpha}\frac{p(g)}{4\lambda_s(g)^2}\mathrm{d}g}_{\text{\rm Quantization Variance}}\nonumber\\
    &~~+ \underbrace{2d\int_{\alpha}^{+\infty}(g-\alpha)^2p(g)\mathrm{d}g}_{\text{\rm Truncation Bias}},
\end{align}
 
Using the Assumption~\ref{ass:stochastic_gradient} and Eq.~\eqref{eq:bias_varance}, we have
\begin{align}\label{eq:error_of_aggregated_g}
    &~~~~~\mathbb{E}[\|\bm{\bar g}_t - \nabla F(\bm{\theta}_t)\|^2]\nonumber\\
    &= \cfrac{\sigma^2}{BN} + \cfrac{d}{N}\int_{-\alpha}^{\alpha}\frac{p(g)}{4\lambda_s(g)^2}\mathrm{d}g + \cfrac{2d}{N}\int_{\alpha}^{+\infty}(g-\alpha)^2p(g)\mathrm{d}g
\end{align}

Assumption~\ref{ass:smoothnesee} further implies that $\forall \bm{\theta},\bm{\theta}' \in \mathbb{R}^d$, we have \begin{equation}
F(\bm{\theta}') \leq F(\bm{\theta}) + \nabla F(\bm{\theta})^\mathrm{T} (\bm{\theta}'-\bm{\theta}) + \frac{\nu}{2} \|\bm{\theta}'-\bm{\theta}\|^2.
\label{eq:smooth}
\end{equation}
Hence, we can get
\begin{align*}
F(\bm{\theta}_{t+1}) &\le F(\bm{\theta}_t) + \nabla F(\bm{\theta}_t)^\text{T} (\bm{\theta}_{t+1}-\bm{\theta}_t) + \cfrac{\nu}{2} \|\bm{\theta}_{t+1}-\bm{\theta}_t\|^2\\
&= F(\bm{\theta}_t) - \eta \nabla F(\bm{\theta}_t)^\top \bm{\bar g}_t + \cfrac{\nu\eta^2}{2} \|\bm{\bar g}_t\|^2\\
&\overset{(a)}{\le} F(\bm{\theta}_k) - \eta \nabla F(\bm{\theta}_t)^\top \bm{\bar g}_t + \cfrac{\eta}{2} \|\bm{\bar g}_t\|^2\\
&= F(\bm{\theta}_t) - \cfrac{\eta}{2} \|\nabla F(\bm{\theta}_t)\|^2 + \cfrac{\eta}{2} \|\bm{\bar g}_t-\nabla F(\bm{\theta}_t)\|^2
\end{align*}
where $(a)$ using $\eta \le \cfrac{1}{\nu}$. Then using Eq.~\eqref{eq:error_of_aggregated_g}, we have
\begin{align*}
    \mathbb{E} F(\bm{\theta}_{t+1}) &\le F(\bm{\theta}_t) - \cfrac{\eta}{2} \|\nabla F(\bm{\theta}_t)\|^2 + \cfrac{\eta}{2NB}\sigma^2\\
    & + \cfrac{d\eta}{2N}\int_{-\alpha}^{\alpha}\frac{p(g)}{4\lambda_s(g)^2}\mathrm{d}g + \cfrac{\eta d}{N}\int_{\alpha}^{+\infty}(g-\alpha)^2p(g)\mathrm{d}g
\end{align*}
Applying it recursively, this yields:
\begin{align*}
    &\mathbb{E}[F(\bm{\theta}_T)-F(\bm{\theta}_0)] \le -\cfrac{\eta }{2}\sum_{t=0}^{T-1}\|\nabla F(\bm{\theta}_t)\|^2
    + \cfrac{T\eta}{2NB}\sigma^2\\
    &~~~~~~~ + \cfrac{dT\eta}{2N}\int_{-\alpha}^{\alpha}\frac{p(g)}{4\lambda_s(g)^2}\mathrm{d}g + \cfrac{T\eta d}{N}\int_{\alpha}^{+\infty}(g-\alpha)^2p(g)\mathrm{d}g
\end{align*}

Considering that $F(\bm{\theta}_T) \ge F(\bm{\theta}^*)$, so:
\begin{align}\label{eq:convergence_UQ2}
&\frac{1}{T}\sum_{t=0}^{T-1} \|\nabla F(\bm{\theta}_t)\|^2 \le \cfrac{2[F(\bm{\theta}_0)-F(\bm{\theta}^*)]}{T\eta}+\cfrac{\sigma^2}{NB}\nonumber\\
&~~~~~ + \cfrac{d}{N}\int_{-\alpha}^{\alpha}\frac{p(g)}{4\lambda_s(g)^2}\mathrm{d}g + \cfrac{2d}{N}\int_{\alpha}^{+\infty}(g-\alpha)^2p(g)\mathrm{d}g
\end{align}

\subsection{Proof of Lemma 3}
\label{proof_lemma3}
The bound for $\|\bm{g}\|_{\infty}^2$ is attained by applying Markov's inequality to $f(\|\bm{g}\|_{\infty}^2) = \exp{[\sqrt{\lambda\|\bm{g}\|_{\infty}^2}]}$. For an arbitrary $\lambda > 0$,
\begin{align*}
    \exp{\{\sqrt{\lambda\mathbb{E}[\|\bm{g}\|_{\infty}^2]}\}} \overset{(a)}{\le} \mathbb{E}[ \exp{\{\sqrt{\lambda\max_j g_j^2}\}}] \le \sum_{j=1}^{d}\mathbb{E}[\exp{\{\sqrt{\lambda} |g_j|\}}]
\end{align*}
where $(a)$ follows from Jensen’s inequality and definition of $\|\cdot\|_{\infty}$. Since $p(g|\gamma) = Laplace(g|0, \gamma)$,
\begin{align*}
    \mathbb{E}[\exp{\{\sqrt{\lambda} |g_j|\}}] = \cfrac{1}{1-\gamma \sqrt{\lambda}}
\end{align*}
Therefore,
\begin{align*}
    \mathbb{E}[\|\bm{g}\|_{\infty}^2] \le \cfrac{1}{\lambda}\ln{\frac{d}{1-\gamma \sqrt{\lambda}}}
\end{align*}
Setting $\sqrt{\lambda} = \frac{1}{2\gamma}$ gives the desired bound in Lemma3.


\subsection{Proof of Theorem 1}
\label{proof_theorem1}
If we set $\alpha = 3\ln{\Big[1+\frac{\sqrt{6}s}{9}\Big]}\gamma$, then the truncated quantization error is
\begin{align*}
   \mathcal{E}_{TQ} = \cfrac{27d\gamma^2}{N(s+\frac{3\sqrt{6}}{2})^2}
\end{align*}

Replacing $\cfrac{d}{N}\int_{-\alpha}^{\alpha}\frac{p(g)}{4\lambda_s(g)^2}\mathrm{d}g + \cfrac{2d}{N}\int_{\alpha}^{+\infty}(g-\alpha)^2p(g)\mathrm{d}g$ with $\mathcal{E}_{TQ}$ in Eq.~\eqref{eq:convergence_UQ2}, then we have
\begin{align*}
    \frac{1}{T}\sum_{t=0}^{T-1} \|\nabla F(\bm{\theta}_t)\|^2 \le \mathcal{E}_{DSGD} + \cfrac{27d\gamma^2}{N(s+\frac{3\sqrt{6}}{2})^2}
\end{align*}



\end{document}